\documentclass[sigconf]{acmart}

\usepackage{amsmath,amssymb,amsfonts}

\usepackage{algorithm}
\usepackage{algpseudocode}
\usepackage{multirow}
\usepackage{threeparttable}
\usepackage{hyperref}

\usepackage{amsthm}
\theoremstyle{plain}

\usepackage{graphicx}
\usepackage{subfigure}
\usepackage{textcomp}
\usepackage{xcolor}

\AtBeginDocument{%
  \providecommand\BibTeX{{%
    \normalfont B\kern-0.5em{\scshape i\kern-0.25em b}\kern-0.8em\TeX}}}

\copyrightyear{2025}
\acmYear{2025}
\acmConference[GECCO '25]{Genetic and Evolutionary Computation Conference}{July 14--18, 2025}{July 14--18, 2025, Malaga, Spain}
\acmBooktitle{Genetic and Evolutionary Computation Conference (GECCO '25), July 14--18, 2025, Malaga, Spain}\acmDOI{10.1145/3712256.3726343}
\acmISBN{979-8-4007-1465-8/2025/07}

\begin{document}



\title[Quality Diversity Genetic Programming for Learning Scheduling Heuristics]{Quality Diversity Genetic Programming\\for Learning Scheduling Heuristics}
\author{Meng Xu}
\email{xu\_meng@simtech.a-star.edu.sg}
\orcid{0000-0002-9930-0403}
\affiliation{
  \institution{Singapore Institute of Manufacturing Technology\\ Agency for Science, Technology and Research (A*STAR)}
  \city{Singapore}
  \country{Republic of Singapore}
}

\author{Frank Neumann}
\orcid{0000-0002-2721-3618}
\affiliation{
  \institution{Optimisation and Logistics\\ School of Computer and Mathematical Sciences\\ The University of Adelaide}
  \city{Adelaide}
  \country{Australia}
}

\author{Aneta Neumann}
\orcid{0000-0002-0036-4782}
\affiliation{
  \institution{Optimisation and Logistics\\ School of Computer and Mathematical Sciences\\ The University of Adelaide}
  \city{Adelaide}
  \country{Australia}
}

\author{Yew Soon Ong}
\email{asysong@ntu.edu.sg}
\orcid{0000-0002-4480-169X}
\affiliation{
  \institution{College of Computing and Data Science, Nanyang
Technological University (NTU)}
\institution{Centre for Frontier AI Research, A*STAR}
  \city{Singapore}
  \country{Republic of Singapore}
}


\renewcommand{\shortauthors}{Meng Xu, Frank Neumann, Aneta Neumann, and Yew Soon Ong}


\begin{abstract}
Real-world optimization often demands diverse, high-quality solutions. Quality-Diversity (QD) optimization is a multifaceted approach in evolutionary algorithms that aims to generate a set of solutions that are both high-performing and diverse. QD algorithms have been successfully applied across various domains, providing robust solutions by exploring diverse behavioral niches. However, their application has primarily focused on static problems, with limited exploration in the context of dynamic combinatorial optimization problems. Furthermore, the theoretical understanding of QD algorithms remains underdeveloped, particularly when applied to learning heuristics instead of directly learning solutions in complex and dynamic combinatorial optimization domains, which introduces additional challenges. This paper introduces a novel QD framework for dynamic scheduling problems. We propose a map-building strategy that visualizes the solution space by linking heuristic genotypes to their behaviors, enabling their representation on a QD map. This map facilitates the discovery and maintenance of diverse scheduling heuristics. Additionally, we conduct experiments on both fixed and dynamically changing training instances to demonstrate how the map evolves and how the distribution of solutions unfolds over time. We also discuss potential future research directions that could enhance the learning process and broaden the applicability of QD algorithms to dynamic combinatorial optimization challenges.
\end{abstract}



\maketitle



\section{Introduction}
Recent research highlights the competitiveness of diversity-seeking optimization approaches, such as Quality-Diversity (QD) \cite{cully2017quality, flageat2023uncertain}. Encouraging the generation of creative and diverse solutions has been shown to enhance exploration \cite{qin2024optimizing}, facilitate the discovery of stepping stones toward novel solutions, and improve adaptability to unforeseen situations. However, current QD methods primarily focus on searching within the solution space, leaving their potential in the heuristic space largely unexplored. This paper addresses these gaps by applying QD to dynamic and large-scale problems, exemplified by the dynamic flexible job shop scheduling (DFJSS) problem. By shifting the focus of QD from directly finding schedule solutions to learning scheduling heuristics, this study significantly broadens the applicability of QD. It opens new pathways for addressing complex, real-world challenges where adaptability and diversity in decision-making strategies are crucial.

DFJSS problem is a complex combinatorial optimization problem that involves scheduling a set of jobs on a set of machines in dynamic environments, where factors such as job arrival times can change over time \cite{shahgholi2019heuristic}. Existing studies in DFJSS typically focus on learning either a single scheduling heuristic \cite{guo2024improved} or an ensemble of scheduling heuristics \cite{xu2023genetic2} to produce a single scheduling solution for a specific DFJSS scenario. However, these approaches lack flexibility, as they do not provide decision-makers with a diverse set of alternative heuristics and corresponding scheduling solutions for comparative analysis and selection—especially in response to unexpected disruptions. In real-world applications, DFJSS problems often extend to supply chain networks \cite{ceylan2021coordinated}, where localized disruptions (e.g., supplier delays) can propagate and severely impact operations. Developing multiple diverse, high-quality scheduling heuristics enables systems to dynamically generate alternative schedules, allowing for real-time adjustments in priorities and resource allocations to mitigate the effects of such disruptions.

This highlights the critical need to study the QD of solutions in DFJSS problems. Focusing on learning multiple high-quality, diverse scheduling heuristics enhances DFJSS systems' adaptability, robustness, and alignment with real-world challenges. Rather than relying on a single heuristic, leveraging a portfolio of diverse heuristics supports dynamic decision-making, better addressing the varied and evolving demands of modern manufacturing and logistics. Additionally, analyzing QD is essential for understanding both the complexity of DFJSS problems and the vastness of their solution spaces, potentially uncovering even more effective individual heuristics. In the context of existing genetic programming (GP) \cite{koza1994genetic} methods for learning scheduling heuristics, QD analysis offers the potential to provide valuable insights into understanding the search space of GP for learning scheduling heuristics—an area that remains largely unexplored. Gaining this deeper understanding is crucial for designing more effective algorithms and heuristics, equipping DFJSS systems to better manage the variability and unpredictability inherent in dynamic environments. However, it is challenging to generate a QD map for individuals (scheduling heuristics) in DFJSS for the following reasons:
\begin{enumerate}
    \item Scheduling heuristics, represented as tree structures, can be mapped to behaviors based on phenotypic characteristics \cite{hildebrandt2015using}. However, the vast space of potential behaviors poses challenges in constructing a QD map. Determining appropriate descriptors to effectively represent and build the map remains an open issue that requires further investigation.
    \item The ever-changing nature of DFJSS problems, with varying job arrival times, makes it challenging to generate a static map that accurately represents the relationship between genotypes and behaviors across all possible scenarios. The context-dependent nature of heuristics means that the effectiveness of a heuristic may vary depending on the specific problem instance.
    \item Changing the training instance(s) for each new generation is a common strategy to enhance the generalization ability of scheduling heuristics in DFJSS. However, this approach complicates the maintenance of a static map and makes it challenging to directly compare individuals across different generations. Consequently, this complexity makes the map update process more difficult and less straightforward.
    \item Evaluating the performance of a heuristic in a DFJSS problem is often computationally expensive, especially for large-scale problems. This makes it difficult to generate a comprehensive map of the heuristic space within a limited time.
\end{enumerate}

In summary, studying QD in DFJSS requires addressing the following challenges step-by-step:
\begin{enumerate}
    \item How to determine descriptors to construct a QD map?
    \item How can the search space be wisely reduced to construct a fixed (static), appropriately sized search space and map?
    \item How can individuals be compared, and how can the map be efficiently updated, especially when individuals are evaluated using different training instances?
\end{enumerate}

This paper proposes a QD map-building strategy integrated with GP for combinatorial optimization, using the DFJSS problem as a case study. This approach provides a tool to visualize the solution space and analyze population diversity, guiding the evolutionary process toward learning high-quality, diverse heuristics.

\section{Preliminaries}
\subsection{Dynamic Flexible Job Shop Scheduling}
In DFJSS \cite{xu2023genetic2}, a set of machines, $M = \{M_1, M_2, \dots, M_m\}$, is available to process a set of jobs, $J = \{J_1, J_2, \dots, J_n\}$. Each job $J_i$ comprises a sequence of operations $[O_{i1}, O_{i2}, \dots, O_{ik}]$, where each operation $O_{ij}$ can be processed on a subset of machines $M_{ij} \subseteq M$, allowing flexibility in machine selection. The capabilities of machines vary, and the processing time of operation $O_{ij}$ on machine $M_m$ is denoted as $p_{ijm}$. Jobs arrive dynamically over time, with $r_i$ representing the release time of job $J_i$ (when the job becomes available for processing), and $d_i$ representing the due date for job $J_i$. The core challenge lies in determining optimal machine assignments and job sequencing while adapting to dynamic, real-time changes in the system. The objective is to optimize key performance metrics, such as minimizing mean flowtime, to enhance overall operational efficiency. This study focuses on a DFJSS problem without considering machine breakdowns or repairs. The problem adheres to the following constraints \cite{xu2023genetic2}: 
\begin{enumerate}
    \item Precedence constraint: Operations in a job must follow a predefined sequence:  
    \begin{equation}
    \footnotesize
        S_{ij} \geq C_{i(j-1)}, \quad \forall j > 1
    \end{equation}
    Here, $S_{ij}$ is the start time of operation $O_{ij}$, and $C_{i(j-1)}$ is the completion time of the preceding operation.
    \item Machine capacity constraint: Each machine can process only one operation at a time:  
    \begin{equation}
    \footnotesize
        S_{ij} \geq C_{uv}, \quad \text{if } x_{ijm} = 1 \text{ and } x_{uvm} = 1, \quad \forall i \neq u, j \neq v
    \end{equation}
    This ensures no overlap of operations on the same machine.
    \item Release time constraint: Operations cannot begin before their job's release time:  
    \begin{equation}
    \footnotesize
        S_{ij} \geq r_i
    \end{equation}
    \item Machine flexibility constraint: Each operation must be assigned to one machine within its eligible set:  
    \begin{equation}
    \footnotesize
        \sum_{m \in M_{ij}} x_{ijm} = 1, \quad \forall O_{ij}
    \end{equation}
\end{enumerate}
This study considers max flowtime ($F_{\text{max}}$), mean tardiness ($T_{\text{mean}}$), and mean weighted flowtime ($WF_{\text{mean}}$) as objectives for the DFJSS problem to evaluate scheduling performance and conduct QD analysis. The metric of $F_{\text{max}}$, $T_{\text{mean}}$, and $WF_{\text{mean}}$ are defined as follows:  
\begin{equation}
\footnotesize
    F_{\text{max}} = \max_{i \in \{1, \dots, n\}} (C_i - r_i), 
\end{equation}
\begin{equation}
\footnotesize
    T_{\text{mean}} = \frac{1}{n} \sum_{i=1}^{n} \max(C_i - d_i, 0), WF_{\text{mean}} = \frac{1}{n} \sum_{i=1}^{n} W_i(C_i - r_i),
\end{equation}
where, $C_i - r_i$ is the flowtime of job $J_i$, calculated as the time elapsed from its release time $r_i$ to its completion time $C_i$. $W_i$ denotes the weight of job $J_i$ and $\max(C_i - d_i, 0)$ represents its tardiness.

\subsection{Quality Diversity Optimization}
QD optimization is a powerful and innovative approach in evolutionary algorithms that focuses on not only finding high-quality solutions but also ensuring diversity within the solution set \cite{cully2017quality}. Traditional optimization techniques typically aim to find a single optimal solution to a problem. However, in many real-world scenarios, especially in complex or dynamic problems, it is important to generate a diverse set of solutions that are both effective and exhibit varied characteristics \cite{nikfarjam2024quality, gounder2024evolutionary}. This enables decision-makers to explore multiple alternatives and choose the solution that best fits their needs, while also preventing overfitting to a single, potentially narrow solution \cite{gounder2024evolutionary}.

The core idea behind QD optimization is to combine two competing objectives: quality (performance) and diversity (variety) \cite{schmidbauer2024guiding,nikfarjam2023evolutionary}. In evolutionary algorithms, this involves exploring diverse solutions while maintaining high quality \cite{nikfarjam2024use}. QD algorithms achieve this by evaluating both a solution's performance and its behavioral characteristics, captured by a behavioral descriptor \cite{goel2024hardening}, thus evolving towards both high performance and behavioral diversity \cite{willemsen2023using}. A key feature is the behavioral space, where solutions are mapped based on behavior, enabling more comprehensive exploration \cite{pugh2016quality}. Common QD algorithms include Novelty Search \cite{kistemaker2011critical, cardoso2021using} and MAP-Elites \cite{cully2021multi}, with successful applications in robotics, control systems, game design, and machine learning.

Despite its successes, QD optimization is still an emerging field, with many open challenges. Dynamic environments present significant challenges for QD algorithms, as they complicate the accurate quantification of both the true performance and the novelty of solutions \cite{flageat2023uncertain}. Furthermore, while QD analysis has traditionally focused on solution space, its application to heuristic space remains largely unexplored. These gaps underscore the need to extend QD methods to effectively address dynamic environments and to tackle problems involving high-dimensional heuristic spaces \cite{anh_viet_do__2024}. GP for DFJSS exemplifies such a domain, involving the use of GP search within heuristic space to learn heuristics for solving complex DFJSS problems. QD has been explored in GP for various applications \cite{boisvert2021quality, zhang2023map}. In \cite{boisvert2021quality}, it is used to evolve decision tree ensembles, optimizing both individual accuracy and collective diversity for classification tasks. Similarly, \cite{zhang2023map} applies QD to GP for learning ensembles in symbolic regression problems.
However, these studies do not focus on the complexity and dynamics of combinatorial optimization problems, where the search space is vast, and constructing a QD map is particularly challenging. Thus, a QD method for effectively addressing complex dynamic combinatorial optimization problems is essential.

\section{QDGP for Learning Heuristics}
\begin{algorithm}[t]
\footnotesize
\caption{QDGP}
\label{GPMAP-Elites}
\begin{algorithmic}[1]
\Require Problem-specific terminal set $T$, function set $F$, population size $P$, maximum evaluations $G$
\Ensure Final QD map $\Lambda $
\State Initialize an empty QD map $\Lambda \gets \emptyset $;
\State Initialize population $P_0$ with $P$ random individuals based on terminal set $T$ and function set $F$;
\State Evaluate fitness of each individual in $P_{0}$;
\State Calculate the behavior of each individual in $P_{0}$;
\State Update individuals to QD map $\Lambda \gets \Lambda \cup P_0$;
\For{$t = 1$ to $P\times G$}
    \State Select parents for breeding based on fitness from QD map $\Lambda $;
    \State Apply crossover/mutation and generate a new individual $ind*$;
    \State Evaluate fitness of the individual $ind*$;
    \State Calculate the behavior of the individual $ind*$;
    \State Update the individual $ind*$ to QD map $\Lambda \gets \Lambda \cup ind*$;
\EndFor
\State Return the QD map $\Lambda $;
\end{algorithmic}
\end{algorithm}

This paper proposes a novel approach that integrates GP with a widely used QD method, MAP-Elites, resulting in the proposed QDGP. As detailed in Algorithm \ref{GPMAP-Elites}, QDGP is utilized to evolve a diverse set of scheduling heuristics (individuals). To represent the behavior of each individual, the method adopts phenotypic characterization (PC) \cite{hildebrandt2015using}. The genotype (individual) is applied to a subset of decision points ($2d$) in an unseen DFJSS instance to derive the PC. These decision points include both sequencing ($d$) and routing ($d$) decisions. At each decision point, we consider $c$ candidate machines or operations ($[0,\cdots,c]$). The value of $d$ and $c$ can be adjusted based on the specific problem and the desired map size. In this study, we concentrate on a specific set of $4$ sequencing decision points and $4$ routing decision points ($d=4$) and $3$ candidates ($c=3$) to prevent the creation of an excessively large map. This approach, using a limited number of decision points to estimate behavior, might not differentiate scheduling heuristics with small behavior differences. As a result, a significant number of individuals might be grouped into the same niche, sharing identical or similar behaviors, which could obscure meaningful distinctions between their behaviors. However, in real-world applications, it is not very useful to get multiple scheduling heuristics with very similar behavior, it is more meaningful to get multiple scheduling heuristics with quite different behaviors. So the above method, using a limited number of decision points to estimate behavior, will put similar scheduling heuristics into the same niche, which is actually a good way for real-world applications. To be continued, at each decision point, the heuristic (represented by the genotype) must select one of $3$ available candidate machines or operations ($[0,1,2]$). Instead of directly using the raw decisions (i.e., the indices of machines or jobs) as the values for characterizing an individual's behavior, we apply a reference rule (i.e., a manually designed heuristic) to rank the candidates at each decision point \cite{zhang2021surrogate}. Specifically, the reference rule calculates candidates' priorities, which are then used to establish their rankings. After obtaining these rankings, we then identify the specific decision that the individual makes at each point. The rank of the machine or job determined by the reference rule, as selected by the individual, is used as the decision for that individual. This approach provides an easy and clear relationship between individuals' behaviors. An illustrative example of calculating the PC for an individual is shown in Table \ref{PCtable}, considering $4$ sequencing decision points and $4$ routing decision points with each decision point considering $3$ candidates based on the reference scheduling heuristic. According to the given description, the PC of this example is a combination of sequencing decisions and routing decisions, which is $[2,0,1,2,2,0,2,1]$. 

To encode these decisions in a compact manner, we employ a 3-base method. This method maps the PC for $3$ decision points of sequencing/routing rule to a unique value ranging from 0 to 80, as there are $3^4=81$ possible combinations. This encoding scheme allows us to represent the heuristic's decisions efficiently. To build a QD map (heatmap), the encoded values for the $4$ sequencing decision points are used as the x-axis values. The encoded values for the $4$ routing decision points are used as the y-axis values. Each heuristic is mapped to a point in the 2D space, representing its specific combination of sequencing and routing decisions. Then, a QD map can be used to store and visualize the fitness (performance) of heuristics in different cells of the space.

\begin{table}[t]
\centering
\footnotesize
\caption{An example of calculating the PC of an individual.}
\label{PCtable}
\vspace{-3mm}
\setlength\tabcolsep{1pt}
\begin{tabular}{cccc|cccc}
\hline
\multicolumn{4}{c|}{\textbf{Sequencing}}                                                                                                                                                              & \multicolumn{4}{c}{\textbf{Routing}}                                                                                                                                                               \\ \hline
\begin{tabular}[c]{@{}c@{}}Decision\\ points\end{tabular} & \begin{tabular}[c]{@{}c@{}}Reference\\ rule\end{tabular} & \begin{tabular}[c]{@{}c@{}}Sequencing\\ rule\end{tabular} & Decision           & \begin{tabular}[c]{@{}c@{}}Decision\\ points\end{tabular} & \begin{tabular}[c]{@{}c@{}}Reference\\ rule\end{tabular} & \begin{tabular}[c]{@{}c@{}}Routing\\ rule\end{tabular} & Decision           \\ \hline
1($O_{0}$)                                                & 0                                                        & 2                                                         & \multirow{3}{*}{2} & 1($M_{0}$)                                                & \underline 2                                             & \underline 0                                           & \multirow{3}{*}{2} \\
1($O_{1}$)                                                & \underline 2                                             & \underline 0                                              &                    & 1($M_{1}$)                                                & 1                                                        & 1                                                      &                    \\
1($O_{2}$)                                                & 1                                                        & 1                                                         &                    & 1($M_{2}$)                                                & 0                                                        & 2                                                      &                    \\ \hline
2($O_{0}$)                                                & 2                                                        & 1                                                         & \multirow{3}{*}{0} & 2($M_{0}$)                                                & 1                                                        & 2                                                      & \multirow{3}{*}{0} \\
2($O_{1}$)                                                & \underline 0                                             & \underline 0                                              &                    & 2($M_{1}$)                                                & 2                                                        & 1                                                      &                    \\
2($O_{2}$)                                                & 1                                                        & 2                                                         &                    & 2($M_{2}$)                                                & \underline 0                                             & \underline 0                                           &                    \\ \hline
3($O_{0}$)                                                & 0                                                        & 1                                                         & \multirow{3}{*}{1} & 3($M_{0}$)                                                & 0                                                        & 2                                                      & \multirow{3}{*}{2} \\
3($O_{1}$)                                                & \underline 1                                             & \underline 0                                              &                    & 3($M_{1}$)                                                & \underline 2                                             & \underline 0                                           &                    \\
3($O_{2}$)                                                & 2                                                        & 2                                                         &                    & 3($M_{2}$)                                                & 1                                                        & 1                                                      &                    \\ \hline
4($O_{0}$)                                                & 1                                                        & 2                                                         & \multirow{3}{*}{2} & 4($M_{0}$)                                                & 2                                                        & 2                                                      & \multirow{3}{*}{1} \\
4($O_{1}$)                                                & \underline 2                                             & \underline 0                                              &                    & 4($M_{1}$)                                                & 0                                                        & 1                                                      &                    \\
4($O_{2}$)                                                & 0                                                        & 1                                                         &                    & 4($M_{2}$)                                                & \underline 1                                             & \underline 0                                           &                    \\ \hline
\end{tabular}
\vspace{-3mm}
\end{table}

This paper introduces two QDGP variants designed to address different training instance availability scenarios: QDGPf, for cases with limited training data (using fixed training instances), and QDGP, for cases with enough data (varying training instances per generation). The core distinction between QDGPf and QDGP lies in the QD map update process.

Updating the QD map in QDGPf is straightforward, involving direct fitness comparisons between individuals (scheduling heuristics). However, this direct comparison is not applicable to QDGP, where each generation uses different training instances. While re-evaluating the entire QD map each generation would enable direct comparisons, it would drastically increase computational cost. To mitigate this, we propose a strategy that obviates the need for re-evaluation while still enabling cross-generational comparisons and continuous QD map updates for QDGP. This strategy employs a manually designed rule as a performance baseline. The manual rule is evaluated on each new instance, and individual fitness is normalized relative to this baseline.

Specifically, let the baseline manual rule be denoted as $r(\cdot)$, and let two scheduling heuristics (individuals) evaluated on two different instances, $ins_1$ and $ins_2$, be represented as $h_1(\cdot)$ and $h_2(\cdot)$, with their respective performance scores $fit_1(ins_1)$ and $fit_2(ins_2)$. The manual rule is evaluated on the same two instances, yielding the results $r(ins_1)$ and $r(ins_2)$. The normalized fitness of the two scheduling heuristics is then calculated as:
\begin{equation}
\footnotesize
\begin{cases}
\text{Normalized fitness of } h_1 = \frac{fit_1(ins_1)}{r(ins_1)}\\
\text{Normalized fitness of } h_2 = \frac{fit_2(ins_2)}{r(ins_2)}
\end{cases}
\end{equation}
To account for the general trend of improvement across generations in evolutionary processes, a penalty factor, $\delta$, is introduced. This factor penalizes individuals from earlier generations. The penalized normalized fitness is calculated as:
\begin{equation}
\footnotesize
\begin{cases}
\text{Penalized normalized fitness of } h^*_1 = \delta^{(g_{c}-g_{1})} \times \frac{fit_1(ins_1)}{r(ins_1)} \\
\text{Penalized normalized fitness of } h^*_2 = \delta^{(g_{c}-g_{2})} \times \frac{fit_2(ins_2)}{r(ins_2)}
\end{cases}
\end{equation}
where $g_{1}$ and $g_{2}$ are the generations in which $h_1(\cdot)$ and $h_2(\cdot)$ were evaluated, respectively, and $g_{c}$ is the current generation. This penalized normalization enables consistent comparison of heuristic performance against a stable baseline, ensuring comparability across generations without requiring re-evaluation and facilitating efficient QD map updates.

\section{Experiment Design}
\label{design}

\subsection{Dataset}
For our experiments, we generate DFJSS datasets using the simulation model described in \cite{xu2023genetic2}. We simulate a job shop with 10 heterogeneous machines processing 1,500 jobs, with the initial 500 jobs serving as a warm-up period. Each machine's processing rate is randomly assigned within the range of 10 to 15 units. Transportation times between machines and between machines and the entry/exit points are sampled from a uniform discrete distribution between 7 and 100 time units. Job arrivals follow a Poisson process, and each job comprises a random number of operations, ranging from 2 to 10, also drawn from a uniform discrete distribution. Job importance varies, with 20\% of jobs assigned a weight of one, 60\% a weight of two, and the remaining 20\% a weight of four. The workload for each operation is randomly selected from a uniform discrete distribution between 100 and 1000 units. A due date factor of 1.5 is applied, setting each job's due date to 1.5 times its total processing time from arrival.

Recognizing the critical role of utilization levels in capturing diverse DFJSS scenarios, we consider two distinct levels: 0.85 and 0.95. These levels represent less and more congested job shop environments, respectively, enabling performance evaluation across varying shop intensities. The specific scenarios examined in this study result from different combinations of utilization levels and objectives. For example, the scenario \textless{}$F_{\text{max}}$, 0.85\textgreater{} represents a DFJSS case with a utilization level of 0.85, optimized for $F_{\text{max}}$.




\subsection{Parameter Setting}
The experiments employ the terminal and function sets detailed in Table \ref{notation} \cite{xu2023genetic2}. The terminal set consists of machine (NIQ, WIQ, MWT), operation (PT, NPT, OWT), job (WKR, NOR, W, TIS, rDD, SL), and transportation (TRANT) features. The function set includes standard two-argument arithmetic operators $(+, -, \times, /, max, min)$, with division by zero protected by returning 1. The $max$ and $min$ functions also operate on two arguments, returning their respective values. Parameter configurations for QDGP and the comparative methods are presented in Table \ref{parameter}. The tournament selection size is initially set to $7$ and adaptively adjusted based on the size of the QD map ($\left | map \right | $) using the formula $7\times \left | map \right | /500$, where $500$ represents the initial population size.

\begin{table}[t]
\caption{The GP terminal and function set for DFJSS.}
\label{notation}
\vspace{-3mm}
\centering
\footnotesize
\begin{threeparttable}
\begin{tabular}{c|l}
\hline
\textbf{Notation} & \multicolumn{1}{c}{\textbf{Description}}                    \\ 
\hline
NIQ               & Number of operations in the queue                       \\
WIQ               & Work in the queue                                       \\
MWT               & Waiting time of the machine = t\tnote{*} - MRT\tnote{*}                  \\
PT                & Processing time of the operation                        \\
NPT               & Median processing time for the next operation                \\
OWT               & Waiting time of the operation = t - ORT\tnote{*}                  \\
WKR               & Work remaining                                          \\
NOR               & Number of operations remaining                           \\
rDD               & Relative due date = DD\tnote{*} - t                           \\
SL             & Slack                           \\
W                 & Job weight                                              \\
TIS               & Time in system = t - releaseTime\tnote{*}                             \\
TRANT             & Transportation time  \\ 
\hline
Function          & $+, -, \times, /, max, min$ \\
\hline
\end{tabular}
\begin{tablenotes}
\item[*] t: current time; MRT: machine ready time; ORT: operation ready time; DD: due date; releaseTime: release time.
\end{tablenotes}
\end{threeparttable}
\vspace{-3mm}
\end{table}

\begin{table}[t]
\caption{The parameter settings for the QDGP method.}
\label{parameter}
\vspace{-3mm}
\centering
\footnotesize
\begin{tabular}{c|l}
\hline
\textbf{Parameter}                   & \multicolumn{1}{c}{\textbf{Value}}                                                          \\ \hline
Population size                      & 500                                                                                         \\
Number of generations                & 200                                                                                         \\
Method for initializing population   & Ramped-half-and-half                                                                        \\
Initial minimum/maximum depth        & 2 / 6                                                                                       \\
Elitism                              & 10                                                                                          \\
Maximal depth                        & 8                                                                                           \\
Crossover rate                       & 0.80                                                                                        \\
Mutation rate                        & 0.15                                                                                        \\
Reproduction rate                    & 0.05                                                                                        \\
Terminal/non-terminal selection rate & 10\% / 90\%                                                                                 \\
Parent selection                     & \begin{tabular}[c]{@{}l@{}}Adaptive tournament selection / \\ Hybird selection\end{tabular} \\ \hline
\end{tabular}
\vspace{-3mm}
\end{table}

\subsection{Comparison Design}
To evaluate the effectiveness of the proposed QDGP approach for DFJSS, we conduct a comparative study against the standard GP baseline. This evaluation explores two distinct training cases: (1) using a fixed set of training instance(s) throughout the evolutionary process; and (2) employing different training instance(s) for each generation. Furthermore, we investigate the influence of different QD parameter settings within our proposed QDGP framework. The following methods are compared when using a fixed set of training instance(s) throughout the evolutionary process:
\begin{itemize}
    \item GPf: Standard GP using fixed training instances.
    \item QDGPf: Proposed QDGP using fixed training instances and adaptive tournament selection.
    \item QDGPf2: Proposed QDGP using fixed training instances, employing a hybrid selection strategy with a 0.5 probability of adaptive tournament selection and a 0.5 probability of random selection.
\end{itemize}

The following methods are compared when employing different training instance(s) for each generation:
\begin{itemize}
    \item GP: Standard GP using different training instance(s) each generation.
    \item QDGP: Proposed QDGP using different training instance(s) each generation, adaptive tournament selection, and a QD map with a capacity of one individual per cell.
    \item QDGP2: Proposed QDGP using different training instance(s) each generation, a hybrid selection strategy (0.5 adaptive tournament selection, 0.5 random selection), and a QD map with a capacity of one individual per cell.
    \item QDGPA5: Proposed QDGP using different training instance(s) each generation, adaptive tournament selection, and a QD map with a capacity of five individuals per cell.
    \item QDGP2A5: Proposed QDGP using different training instance(s) each generation, a hybrid selection strategy (0.5 adaptive tournament selection, 0.5 random selection), and a QD map with a capacity of five individuals per cell.
\end{itemize}

The experiments are designed with the following goals: 1) Compare QDGP variants against their respective GP baselines (GPf, GP) to quantify the benefits of integrating QD search into the GP framework and assess its effect on the quality of evolved scheduling heuristics. 2) Compare methods using adaptive tournament selection (QDGPf, QDGP, QDGPA5) with those using a hybrid selection strategy (QDGPf2, QDGP2, QDGP2A5) to assess selection pressure's impact on QDGP performance and its exploitation-exploration balance. 3) Analyze QD map coverage percentages of all methods to assess QD's impact on search space exploration. 4) Visualize and compare QD maps from GP and QDGP variants to assess QD's impact on heuristic diversity and performance.


\section{Experimental Results}
\label{results}

\subsection{Test Performance}
\label{sub1:testperformance}
Tables \ref{testperformance1} and \ref{testperformance2} present the results of the proposed QDGP methods compared to baseline methods across six DFJSS scenarios using fixed training instance(s) and using varying training instance(s), respectively. Statistical comparisons are conducted using the Wilcoxon test \cite{zimmerman1993relative}. Significant differences are indicated as follows: ``$\uparrow$'' denotes a significantly better result, ``$\downarrow$'' indicates a significantly worse result, and ``='' signifies no significant difference.

\begin{table}[t]
\centering
\caption{The test performance of the proposed QDGP methods and the comparison method across six DFJSS scenarios using fixed training instance(s).}
\vspace{-3mm}
\label{testperformance1}
\footnotesize
\begin{tabular}{c|ccc}
\hline
Scenario                              & GPf & QDGPf & QDGPf2 \\ \hline
\textless{}$F_{\text{max}}$, 0.85\textgreater{}   & 1404.36(65.35) & 1340.16(36.07)($\uparrow$) & 1348.93(32.96)($\uparrow$)(=)        \\
\textless{}$F_{\text{max}}$, 0.95\textgreater{}   & 1489.12(73.61) & 1427.23(60.33)($\uparrow$) & 1435.68(72.91)($\uparrow$)(=)        \\
\textless{}$T_{\text{mean}}$, 0.85\textgreater{}  & 205.17(5.28) & 200.88(3.57)($\uparrow$) & 202.04(5.06)($\uparrow$)(=)        \\
\textless{}$T_{\text{mean}}$, 0.95\textgreater{}  & 262.69(5.05) & 255.77(5.20)($\uparrow$) & 258.15(5.58)($\uparrow$)($\downarrow$)        \\
\textless{}$WF_{\text{mean}}$, 0.85\textgreater{} & 1244.82(13.38) & 1228.08(8.50)($\uparrow$) & 1232.62(12.20)($\uparrow$)(=)        \\
\textless{}$WF_{\text{mean}}$, 0.95\textgreater{} & 1355.89(16.63) & 1340.73(7.45)($\uparrow$) & 1339.81(6.67)($\uparrow$)(=)        \\ \hline
\end{tabular}
\vspace{-3mm}
\end{table}

\subsubsection{Using Fixed Training Instance(s)}
The results presented in Table \ref{testperformance1} demonstrate the performance of the proposed QDGP methods (QDGPf and QDGPf2) compared to a standard GP baseline (GPf) across six DFJSS scenarios using fixed training instance(s). Statistical comparisons are performed successively from left to right across the table. For instance, 1348.93(32.96)($\uparrow$)(=), in the first row, indicates that QDGPf2 performs significantly better than GP and shows no significant difference compared to QDGPf.

Both proposed QDGP methods (QDGPf and QDGPf2) consistently outperform the GPf across all six scenarios. This indicates that incorporating QD principles into the GP framework leads to significant improvements in the quality of the evolved scheduling heuristics when using a fixed training set. This improvement is observed across different optimization objectives ($F_{\text{max}}$, $T_{\text{mean}}$, $WF_{\text{mean}}$) and utilization levels (0.85 and 0.95), suggesting the robustness of the QD approach in this context. Comparing QDGPf and QDGPf2, we observe that QDGPf generally achieves slightly better or comparable performance than QDGPf2. This suggests that the adaptive tournament selection strategy used in QDGPf is effective in guiding the search towards high-quality solutions within the QD map. The hybrid selection strategy used in QDGPf2, which combines adaptive tournament selection with random selection, does not consistently provide further improvements and in one case even leads to a statistically significant degradation compared to QDGPf ($T_{\text{mean}}$ at utilization 0.95). This indicates that introducing a degree of randomness in parent selection, while potentially promoting diversity, might also hinder the exploitation of promising areas in the search space when using a fixed training set. In most cases, the hybrid selection does not cause a significant difference in performance.

\begin{table*}[t]
\centering
\caption{The test performance of the proposed QDGP methods and the comparison method across six DFJSS scenarios using varying training instance(s) each generation.}
\vspace{-3mm}
\label{testperformance2}
\footnotesize
\begin{tabular}{c|cccll}
\hline
Scenario                              & GP & QDGP & QDGP2 & QDGPA5 & QDGP2A5 \\ \hline
\textless{}$F_{\text{max}}$, 0.85\textgreater{}   & 1298.90(19.77) & 1323.49(19.52)($\downarrow$) & 1360.12(31.24)($\downarrow$)($\downarrow$) & 1329.56(30.46)($\downarrow$)(=)($\uparrow$) & 1340.16(33.00)($\downarrow$)(=)($\uparrow$)(=)         \\
\textless{}$F_{\text{max}}$, 0.95\textgreater{}   & 1381.22(22.06) & 1405.32(22.93)($\downarrow$) & 1452.38(40.67)($\downarrow$)($\downarrow$) & 1422.86(49.49)($\downarrow$)(=)($\uparrow$) & 1441.92(38.24)($\downarrow$)($\uparrow$)(=)($\downarrow$)         \\
\textless{}$T_{\text{mean}}$, 0.85\textgreater{}  & 198.03(2.36) & 202.08(2.88)($\downarrow$) & 202.71(2.96)($\downarrow$)(=) & 200.59(4.20)($\downarrow$)($\uparrow$)($\uparrow$) & 203.22(2.91)($\downarrow$)(=)(=)($\downarrow$)         \\
\textless{}$T_{\text{mean}}$, 0.95\textgreater{}  & 255.05(4.33) & 257.35(4.34)($\downarrow$) & 258.43(4.48)($\downarrow$)(=) & 258.43(5.25)($\downarrow$)(=)(=) & 260.29(5.29)($\downarrow$)(=)(=)(=)         \\
\textless{}$WF_{\text{mean}}$, 0.85\textgreater{} & 1219.36(8.06) & 1231.01(6.30)($\downarrow$) & 1234.60(7.90)($\downarrow$)($\downarrow$) & 1228.42(6.96)($\downarrow$)(=)($\uparrow$) & 1233.30(10.39)($\downarrow$)(=)(=)(=)         \\
\textless{}$WF_{\text{mean}}$, 0.95\textgreater{} & 1332.50(8.50) & 1344.57(10.32)($\downarrow$) & 1350.84(11.43)($\downarrow$)($\downarrow$) & 1346.18(11.16)($\downarrow$)(=)(=) & 1346.77(9.37)($\downarrow$)(=)(=)(=)         \\ \hline
\end{tabular}
\vspace{-3mm}
\end{table*}

\subsubsection{Using Varying Training Instances Each Generation}
Table \ref{testperformance2} compares the test performance of the proposed QDGP methods (QDGP, QDGP2, QDGPA5, and QDGP2A5) against the GP across six DFJSS scenarios using varying training instances each generation. Similarly, statistical comparisons are conducted left to right, as in Table \ref{testperformance1}.


Unlike the results obtained with fixed training instances (Table \ref{testperformance1}), the QDGP variants (QDGP, QDGP2, QDGPA5, and QDGP2A5) generally underperform the GP baseline across most scenarios when using varying training instances each generation. This indicates that directly applying QD in conjunction with dynamic training can hinder performance, suggesting a negative interaction between the dynamic training instances and the basic QD exploration strategy. A possible explanation is that the varying training instances already enhance population diversity, and the additional use of a QD map may disrupt the balance between diversity and convergence.

Comparing QDGP2 to QDGP and QDGP2A5 to QDGPA5 reveals that the hybrid parent selection strategy (50\% adaptive tournament selection, 50\% random selection) did not yield consistent performance improvements. While some instances show no significant difference between variants with and without hybrid selection, the majority exhibit further performance degradation with the hybrid strategy. This suggests that the random selection component exacerbates the disruption to the search process caused by the dynamic training instances. Similarly, increasing the QD map capacity (from one individual per cell in QDGP and QDGP2 to five in QDGPA5 and QDGP2A5) consistently enhances performance. However, significant improvements are observed in only a few instances with larger map sizes. This suggests that merely increasing the QD map capacity is not sufficient to fully counteract the negative effects of dynamic training on the QD search.

These performance degradations are observed across all three objectives ($F_{\text{max}}$, $T_{\text{mean}}$, $WF_{\text{mean}}$) and both utilization levels (0.85 and 0.95), indicating that the adverse interaction between QD and dynamic training is a general trend rather than a scenario-specific anomaly. The observed results suggest that the current QDGP formulation is not well-suited for dynamic training. The frequent changes in training instances likely disrupt the formation and maintenance of niches within the QD map, leading to unstable and less effective exploration of the solution space. Future research should focus on adapting QDGP for dynamic training scenarios, potentially through better strategies such as dynamically adjusting the QD map based on changes in training instances, incorporating better memory mechanisms to retain useful information across generations, or exploring better strategies to balance between diversity and convergence under different training conditions.

Overall, the results demonstrate that while QDGP provides significant benefits with fixed training instances, its direct application with varying training instances each generation leads to performance degradation compared to standard GP. This highlights the need for further research to develop effective QDGP methods for dynamic training scenarios in dynamic scheduling problems.

\subsection{QD Map Coverage Percentage}

\begin{figure}[t]
\centerline{\includegraphics[width=0.45\textwidth]{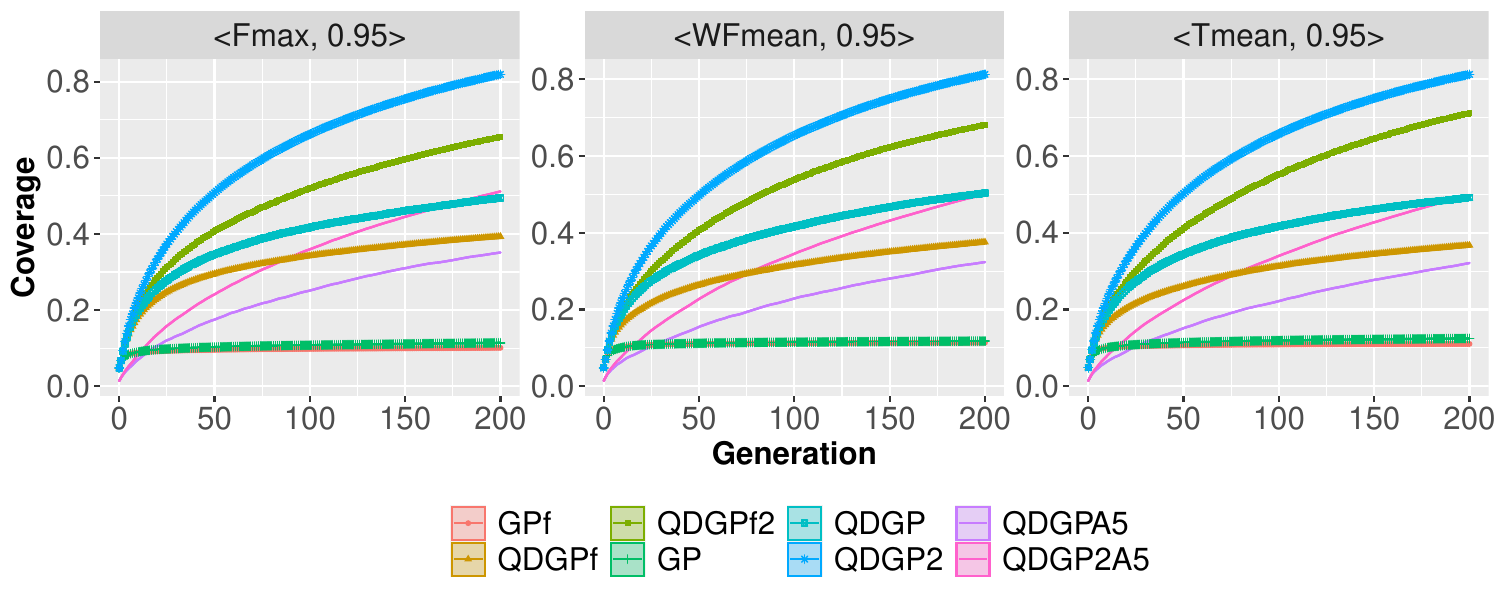}}
\vspace{-3mm}
\caption{The convergence curves of coverage percentages for the examined methods across 3 DFJSS scenarios.}
\label{coverage}
\vspace{-3mm}
\end{figure}

Figure \ref{coverage} illustrates the convergence curves of the coverage percentage for the proposed and baseline methods across three DFJSS scenarios. All methods demonstrate an increasing trend in coverage over generations, indicating exploration of the search space and the discovery of diverse scheduling heuristics, each offering distinct scheduling solutions. QDGP2 generally exhibits the fastest convergence and achieves the highest coverage among all methods. The QDGP variants (QDGP, QDGP2, QDGPA5, and QDGP2A5) consistently outperform the GPf and GP baselines in coverage percentage, demonstrating the benefits of incorporating QD. The hybrid selection strategy (used in QDGP2 and QDGP2A5) tends to accelerate convergence compared to standard selection (used in QDGP and QDGPA5). Increasing the QD map capacity (as in QDGPA5 and QDGP2A5) generally leads to a larger number of discovered scheduling heuristics; however, it does not necessarily translate to proportionally higher coverage relative to the increased map size. Furthermore, the convergence rate of most methods slows in later generations, suggesting diminishing returns in finding new solutions as the search progresses, likely due to the increasing difficulty of exploration.

Integrating the coverage results with the test performance analysis presented in Subsection \ref{sub1:testperformance}, we observe a correlation between higher coverage (indicating a more diverse population) and the potential to learn improved scheduling heuristics. However, excessively high coverage may hinder the discovery of truly high-quality solutions, possibly by overemphasizing exploration at the expense of exploitation. The choice of method appears to depend on the availability of training data. With limited instances (using fixed training instances), QDGP variants, especially QDGPf as in Table \ref{testperformance1}, demonstrate superior performance. Conversely, with sufficient instances (varying training instances each generation), the standard GP (GPf) tends to outperform the QDGP variants, as seen in Table \ref{testperformance2}. Future research could investigate alternative QD strategies aimed at simultaneously enhancing both coverage percentage and the quality of the resulting scheduling heuristics. 


\subsection{QD Map Visulization}
\begin{figure}[t]
	\subfigure[QD maps for GPf.]{
			\centering
			\includegraphics[width=0.44\textwidth]{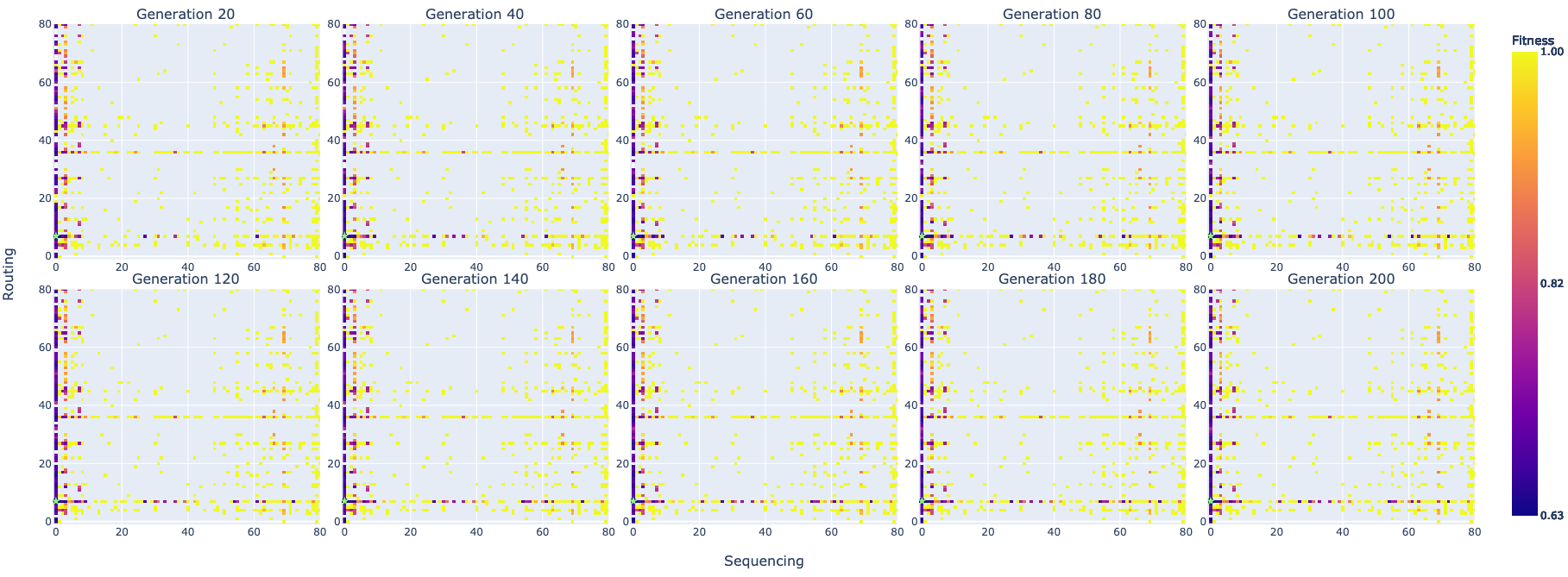}
			\label{figheatmapGPf}
	}\\
    \vspace{-3mm}
	\subfigure[QD maps for QDGPf.]{
			\centering
			\includegraphics[width=0.44\textwidth]{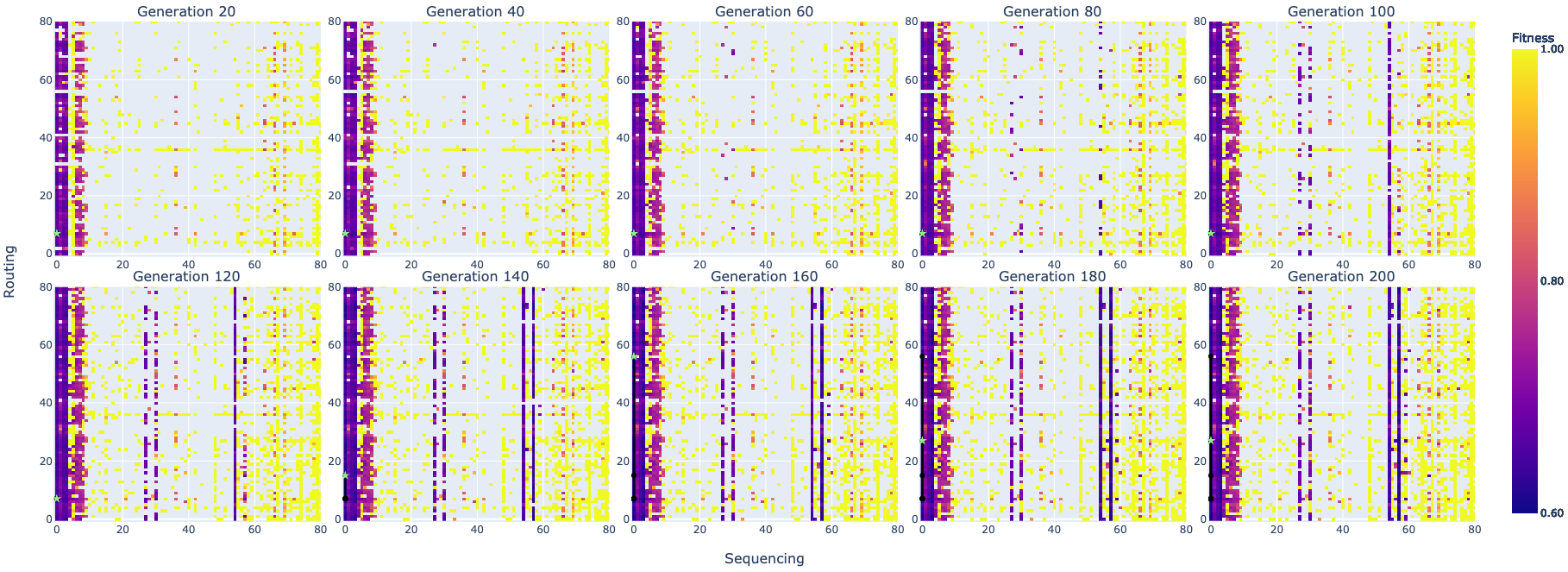}
			\label{figheatmapQDGPf}
	}\\
    \vspace{-3mm}
	\subfigure[QD maps for QDGPf2.]{
			\centering
			\includegraphics[width=0.44\textwidth]{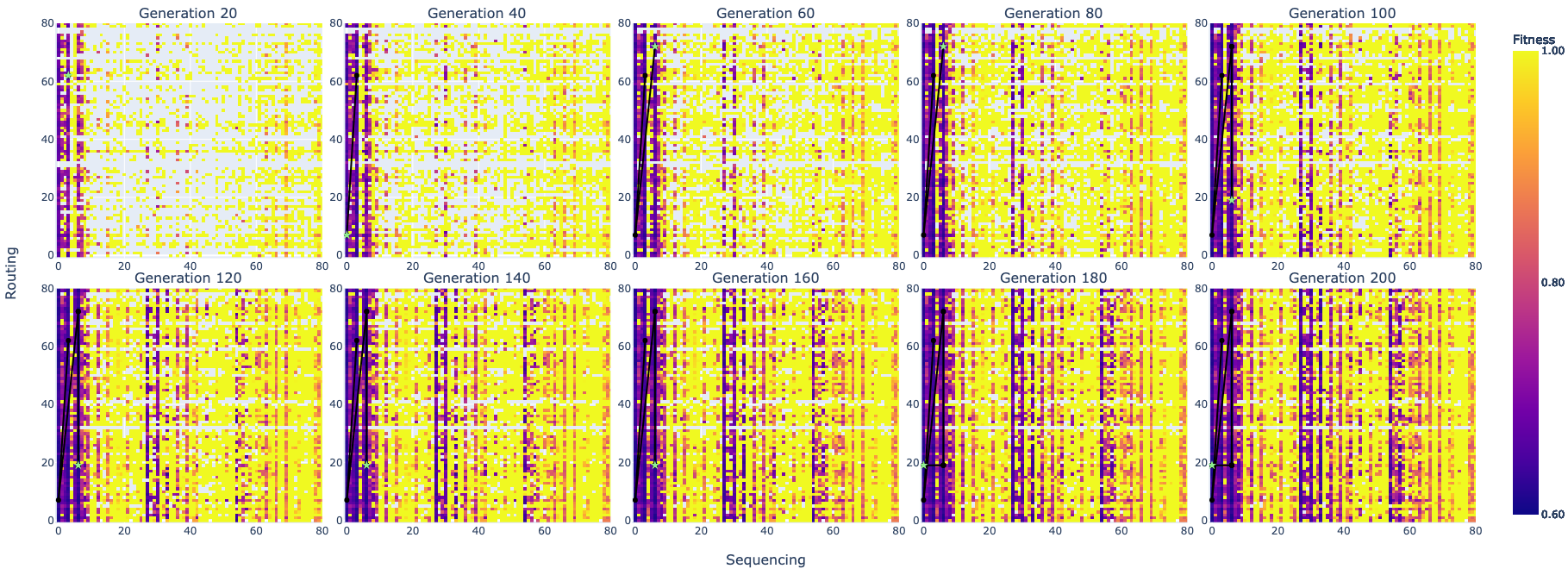}
			\label{figheatmapQDGPf2}
	}%
    \vspace{-3mm}
	\centering
	\caption{Examples of the QD maps that visualize the population distribution at every 20th generation during the evolutionary process of a single run of GPf, QDGPf, and QDGPf2.}
	\label{figheatmap:main}
	\vspace{-5mm}
\end{figure}

QD maps, recorded every 500 evaluations (defined as a generation), are visualized as heatmaps. Figures \ref{figheatmapGPf}, \ref{figheatmapQDGPf}, and \ref{figheatmapQDGPf2} present these maps for GPf, QDGPf, and QDGPf2, respectively, generated using the same DFJSS and fixed training instance(s) each generation. These visualizations demonstrate QD's ability to enhance diversity and discover high-quality scheduling heuristics. In these figures, the green star marks the best solution found, and the black curve traces the trajectory of the best solution over time. To improve visual clarity, fitness values exceeding 1.1 times the best value within each map are capped at this threshold.

These figures illustrate the explored search space of scheduling heuristics. Initially, all three methods exhibit limited QD map coverage, indicating that their initial populations explore only a small portion of the potential solution space. As evolution progresses, however, QDGPf and, especially, QDGPf2 demonstrate significantly enhanced exploration, populating a much larger portion of the QD map with diverse scheduling heuristics. In contrast, the QD map by GPf remains largely unexplored, reflecting a lack of diversity. This highlights the positive impact of QD on search space exploration, with the hybrid selection strategy in QDGPf2 further enhancing this exploration. The proposed behavior-based mapping allows the algorithms to actively seek diverse solutions, mitigating premature convergence and increasing the likelihood of discovering globally competitive or novel scheduling strategies. This is visually evident in the QD maps, where QDGPf and QDGPf2 populate distinct regions, while GPf concentrates its search within a few limited areas. Further, in Figure \ref{figheatmap:main}, starting with the same initial population, GPf, QDGPf, and QDGPf2 exhibit distinct evolutionary trajectories, leading to different scheduling performances and behaviors. GPf achieves the highest performance with a fitness of 0.633, utilizing sequencing behavior 0 and routing behavior 7. In contrast, QDGPf finds its best solution with a fitness of 0.604, employing sequencing behavior 0 and routing behavior 27. QDGPf2, with its hybrid selection strategy, achieves a fitness of 0.605, utilizing sequencing behavior 0 and routing behavior 19. The observation that all three methods identify sequencing behavior 0 as part of their best-performing solutions suggests that this particular sequencing strategy holds intrinsic value within this specific instance. Despite the shared preference for sequencing behavior 0, each method converges on distinct optimal routing behaviors. GPf favors routing behavior 7, while QDGPf and QDGPf2 select routing behaviors 27 and 19, respectively. This highlights the search strategy's role in solution space exploration. QD-based methods, with their emphasis on diversity, explore a wider range of routing options, leading to the discovery of different high-performing combinations.

\begin{table*}[t]
\centering
\footnotesize
\caption{The mean and standard deviation of coverage percentage, mean $T_{\text{mean}}$ of whole QD map, and best $T_{\text{mean}}$ every 20 generations of 30 runs of GPf and our proposed QDGPf and QDGPf2 on the training scenario \textless{}$T_{\text{mean}}$, 0.85\textgreater{}.}
\label{tableall}
\vspace{-3mm}
\begin{tabular}{c|ccc|ccc|ccc}
\hline
\multirow{2}{*}{Generation} & \multicolumn{3}{c|}{Coverage (\%)} & \multicolumn{3}{c|}{Mean $T_{\text{mean}}$} & \multicolumn{3}{c}{Best $T_{\text{mean}}$} \\ \cline{2-10} 
                            & GPf            & QDGPf      & QDGPf2   & GPf            & QDGPf      & QDGPf2     & GPf            & QDGPf    & QDGPf2        \\ \hline
20 & 10.68(0.88) & 19.78(1.59)($\uparrow$) & 26.59(1.62)($\uparrow$)($\uparrow$) & 1.524(0.092) & 1.216(0.094)($\uparrow$) & 1.395(0.097)($\uparrow$)($\downarrow$) & 0.576(0.011) & 0.567(0.013)($\uparrow$) & 0.569(0.014)(=)(=) \\
40 & 10.93(0.88) & 23.59(2.41)($\uparrow$) & 36.67(2.28)($\uparrow$)($\uparrow$) & 1.511(0.092) & 1.205(0.110)($\uparrow$) & 1.414(0.111)($\uparrow$)($\downarrow$) & 0.569(0.013) & 0.560(0.013)($\uparrow$) & 0.559(0.011)($\uparrow$)(=) \\
60 & 11.06(0.87) & 26.26(3.07)($\uparrow$) & 44.12(2.32)($\uparrow$)($\uparrow$) & 1.503(0.090) & 1.220(0.125)($\uparrow$) & 1.411(0.121)($\uparrow$)($\downarrow$) & 0.565(0.012) & 0.556(0.014)($\uparrow$) & 0.555(0.011)($\uparrow$)(=) \\
80 & 11.15(0.88) & 28.37(3.55)($\uparrow$) & 50.26(2.38)($\uparrow$)($\uparrow$) & 1.499(0.093) & 1.224(0.131)($\uparrow$) & 1.398(0.123)($\uparrow$)($\downarrow$) & 0.563(0.012) & 0.553(0.012)($\uparrow$) & 0.554(0.011)($\uparrow$)(=) \\
100 & 11.22(0.88) & 30.12(3.89)($\uparrow$) & 55.34(2.37)($\uparrow$)($\uparrow$) & 1.491(0.093) & 1.223(0.135)($\uparrow$) & 1.379(0.116)($\uparrow$)($\downarrow$) & 0.562(0.012) & 0.551(0.009)($\uparrow$) & 0.553(0.011)($\uparrow$)(=) \\
120 & 11.27(0.87) & 31.62(4.12)($\uparrow$) & 59.75(2.24)($\uparrow$)($\uparrow$) & 1.486(0.097) & 1.229(0.135)($\uparrow$) & 1.361(0.115)($\uparrow$)($\downarrow$) & 0.561(0.011) & 0.547(0.007)($\uparrow$) & 0.551(0.010)($\uparrow$)(=) \\
140 & 11.32(0.91) & 32.95(4.33)($\uparrow$) & 63.50(2.26)($\uparrow$)($\uparrow$) & 1.482(0.092) & 1.231(0.139)($\uparrow$) & 1.342(0.113)($\uparrow$)($\downarrow$) & 0.561(0.011) & 0.547(0.007)($\uparrow$) & 0.549(0.009)($\uparrow$)(=) \\
160 & 11.35(0.92) & 34.12(4.55)($\uparrow$) & 66.83(2.31)($\uparrow$)($\uparrow$) & 1.478(0.092) & 1.232(0.138)($\uparrow$) & 1.318(0.112)($\uparrow$)($\downarrow$) & 0.560(0.012) & 0.545(0.007)($\uparrow$) & 0.549(0.009)($\uparrow$)(=) \\
180 & 11.41(0.97) & 35.19(4.68)($\uparrow$) & 69.70(2.30)($\uparrow$)($\uparrow$) & 1.475(0.091) & 1.234(0.139)($\uparrow$) & 1.294(0.106)($\uparrow$)($\downarrow$) & 0.558(0.012) & 0.545(0.007)($\uparrow$) & 0.548(0.009)($\uparrow$)(=) \\
200 & 11.44(1.00) & 36.12(4.78)($\uparrow$) & 72.25(2.29)($\uparrow$)($\uparrow$) & 1.473(0.089) & 1.241(0.141)($\uparrow$) & 1.273(0.103)($\uparrow$)(=) & 0.557(0.012) & 0.545(0.007)($\uparrow$) & 0.548(0.009)($\uparrow$)(=) \\ \hline
\end{tabular}
\end{table*}

Table \ref{tableall} summarizes the mean and standard deviation of the coverage percentage, mean fitness across the QD map, and best fitness values, recorded every 20 generations across 30 runs for GPf, QDGPf, and QDGPf2 on the \textless{}$T_{\text{mean}}$, 0.85\textgreater{} scenario. Early in the evolutionary process (generation 20), QDGPf achieves a coverage of 19.73\% (1.60\%), significantly surpassing GPf's 10.65\% (0.89\%). Notably, QDGPf2 demonstrates an even higher coverage of 26.67\% (1.59\%), indicating that both QDGPf and QDGPf2 explore a substantially larger portion of the search space, reflecting greater initial diversity compared to GPf. This trend continues throughout the evolutionary process, with the coverage gap between QDGPf/QDGPf2 and GPf progressively widening. By generation 200, QDGPf achieves a coverage of 36.00\% (4.82\%), while QDGPf2 reaches an impressive 72.31\% (2.31\%). In contrast, GPf plateaus at 11.43\% (1.02\%). These results underscore the superior capability of QDGPf and QDGPf2 in diversifying the search and exploring a broader range of scheduling heuristics, with QDGPf2 demonstrating a clear advantage over QDGPf.

Regarding $T_{\text{mean}}$ values, QDGPf and QDGPf2 consistently achieve smaller mean $T_{\text{mean}}$ and best $T_{\text{mean}}$ values compared to GPf as evolution progresses. By generation 200, QDGPf2 achieves the lowest mean $T_{\text{mean}}$ (1.268 $\pm$ 0.102) and best $T_{\text{mean}}$ (0.548 $\pm$ 0.009), followed by QDGPf (mean $T_{\text{mean}}$: 1.244 $\pm$ 0.143, best $T_{\text{mean}}$: 0.545 $\pm$ 0.007), while GPf exhibits higher values (mean $T_{\text{mean}}$: 1.475 $\pm$ 0.090, best $T_{\text{mean}}$: 0.557 $\pm$ 0.012). This indicates that QDGPf and QDGPf2 are not only more effective at exploring the search space but also more efficient in identifying higher-quality solutions, with QDGPf2 demonstrating the most robust performance overall.

\begin{figure}[t]
    \centering
    \includegraphics[width=0.45\textwidth]{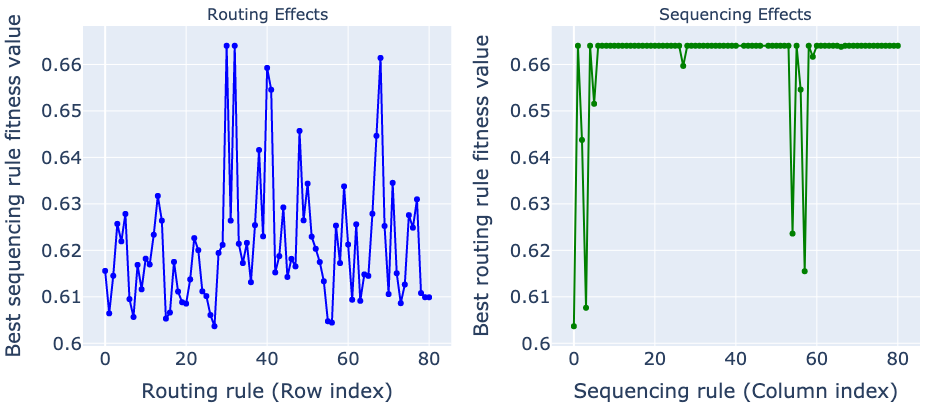}  
    \vspace{-3mm}
    \caption{An example of the visualization identifies the best sequencing/routing rule for each corresponding routing/sequencing rule, as denoted in Figure \ref{figheatmapQDGPf} QD map. }
    \label{figx_y_best_effects}
    \vspace{-3mm}
\end{figure}

To understand the independent contributions of sequencing and routing rules to scheduling performance, we analyze the marginal distributions of fitness values along the x-axis (sequencing) and y-axis (routing) of the final QD map (generation 200). This analysis isolates the effect of each rule type. Figure \ref{figx_y_best_effects} visualizes these marginal effects based on the QD map shown in Figure \ref{figheatmapQDGPf}. The \emph{Routing Effects} subplot displays the best fitness value found within each row of the QD map, representing the optimal fitness achieved for each routing behavior in conjunction with its corresponding optimal sequencing behavior. Conversely, the \emph{Sequencing Effects} subplot shows the best fitness value within each column, representing the optimal fitness for each sequencing behavior given its corresponding optimal routing behavior.

As shown in Figure \ref{figx_y_best_effects}, routing behavior 27 achieves the overall best fitness when paired with its optimal sequencing counterpart. Similarly, sequencing behavior 0 consistently yields the best fitness when combined with its optimal routing counterpart. While a small subset of sequencing behaviors (0, 3, and 57) appear to be dominant, the impact of sequencing variations is less pronounced than that of routing behaviors. Several distinct routing behaviors (e.g., 1, 7, 15, 27, 56) support strong performance, with multiple routing behaviors achieving comparable and near-optimal fitness values. This highlights the crucial role of routing behavior optimization in achieving effective scheduling outcomes. These findings emphasize the importance of prioritizing the evolution and refinement of routing rules to further enhance scheduling performance.



In summary, the evolutionary process effectively explores and refines the search space, diversifies the population, and achieves significant improvements in both mean performance and best fitness values. The combination of visual QD maps and quantitative metrics highlights the method's strength in identifying diverse and high-performing scheduling heuristics, with implications for further optimization efforts.

\section{Conclusions}
\label{conclusion}
The ability to generate diverse sets of high-quality solutions is crucial in evolutionary computation. However, QD research has primarily focused on static, small-scale problems, neglecting the complexities of dynamic, large-scale scenarios. These dynamic problems demand more sophisticated exploration strategies. Moreover, while QD has been extensively studied in the context of solution space search, its application to heuristic space search, as performed by GP, remains relatively unexplored, presenting unique challenges.

To address these gaps, this paper proposes a novel QDGP method with a map-building strategy specifically designed for dynamic combinatorial optimization problems, like DFJSS. This approach shifts the search from the solution space to the heuristic space, enabling a more comprehensive exploration of diverse heuristics adaptable to varying problem instances. A novel map update strategy is also introduced to accommodate dynamically changing training instances, ensuring consistent QD map evolution. Results demonstrate the promise of this approach for large-scale dynamic combinatorial optimization, particularly in data-scarce scenarios, by generating heuristics with diverse behaviors. However, performance with abundant training data is suboptimal, motivating further research into QD analysis in GP for handling dynamic combinatorial optimization problems with abundant training data.

Overall, this work extends the application of QD techniques to heuristic space search, offering new insights into the role of diverse heuristics in dynamic, large-scale optimization. A promising future direction is to explore an adaptive map updating approach, where the map dynamically expands or contracts based on the evolution process. Additionally, this study lays the foundation for future research on QD analysis in GP and other heuristic-space search algorithms to tackle complex, real-world optimization challenges.

\section*{Acknowledgments}
This research is partially supported by the Distributed Smart Value Chain programme, which is funded in part by the Singapore RIE2025 Manufacturing, Trade and Connectivity (MTC) Industry Alignment Fund-Pre-Positioning (Award No: M23L4a0001) and partially supported by the National Research Foundation, Singapore under its AI Singapore Programme (AISG Award No: AISG3-RP-2022-031). This work has also been supported by the Australian Research Council through grants DP190103894 and FT200100536.


%






\newpage



%



\balance
\bibliographystyle{ACM-Reference-Format}


%








\end{document}